\pdfoutput=1

\documentclass[11pt]{article}

\usepackage[]{ACL2023}

\usepackage{times}
\usepackage{latexsym}

\usepackage[T1]{fontenc}

\usepackage[utf8]{inputenc}

\usepackage{microtype}

\usepackage{inconsolata}

\usepackage{url}
\usepackage{times}
\usepackage{latexsym}
\usepackage{bm}
\usepackage{hyperref}
\usepackage{color,colortbl}
\usepackage{makecell}
\usepackage{microtype}
\usepackage{graphicx}
\usepackage{multirow}
\usepackage{subfigure}
\usepackage{booktabs}
\usepackage{colortbl}
\usepackage[T1]{fontenc}
\usepackage[utf8]{inputenc}
\usepackage{todonotes}
\usepackage{xcolor}
\newcommand\blfootnote[1]{%
  \begingroup
  \renewcommand\thefootnote{}\footnote{#1}%
  \addtocounter{footnote}{-1}%
  \endgroup
}

\title{Pre-Trained Language-Meaning Models for\\ Multilingual Parsing and Generation}

\author{
Chunliu Wang$^{*}$, Huiyuan Lai$^{*}$, Malvina Nissim, Johan Bos\\
CLCG, University of Groningen / The Netherlands\\
\texttt{\{chunliu.wang, h.lai, m.nissim, johan.bos\}@rug.nl}
}

\begin{document}

\maketitle

\begin{abstract}

Pre-trained language models (PLMs) have achieved great success in NLP and have recently been used for tasks in computational semantics. 
However, these tasks do not fully benefit from PLMs since meaning representations are not explicitly included in the pre-training stage.
We introduce \textit{multilingual pre-trained language-meaning models} based on Discourse Representation Structures (DRSs), including meaning representations besides natural language texts in the same model, and design a new strategy to reduce the gap between the pre-training and fine-tuning objectives. 
Since DRSs are language neutral, cross-lingual transfer learning is adopted to further improve the performance of non-English tasks.
Automatic evaluation results show that our approach achieves the best performance on both the multilingual DRS parsing and DRS-to-text generation tasks.
Correlation analysis between automatic metrics and human judgements on the generation task further validates the effectiveness of our model. Human inspection reveals that out-of-vocabulary tokens are the main cause of erroneous results.\blfootnote{$^*$ Equal contribution.}
\end{abstract}

\section{Introduction}

There are two common tasks in computational semantics: mapping a text to a meaning representation (semantic parsing), and its reverse, producing a text from a meaning representation (semantic generation). These tasks generally rely on corpora that contain texts aligned with meaning representations.
While in recent years large pre-trained language models (PLMs), both monolingual as well as multilingual, have brought NLP tasks to a new level, semantic parsing and generation cannot fully benefit from them since the meaning representations are not included in PLMs explicitly. 
\begin{figure}[htbp]
\centering
\includegraphics[scale=.51]{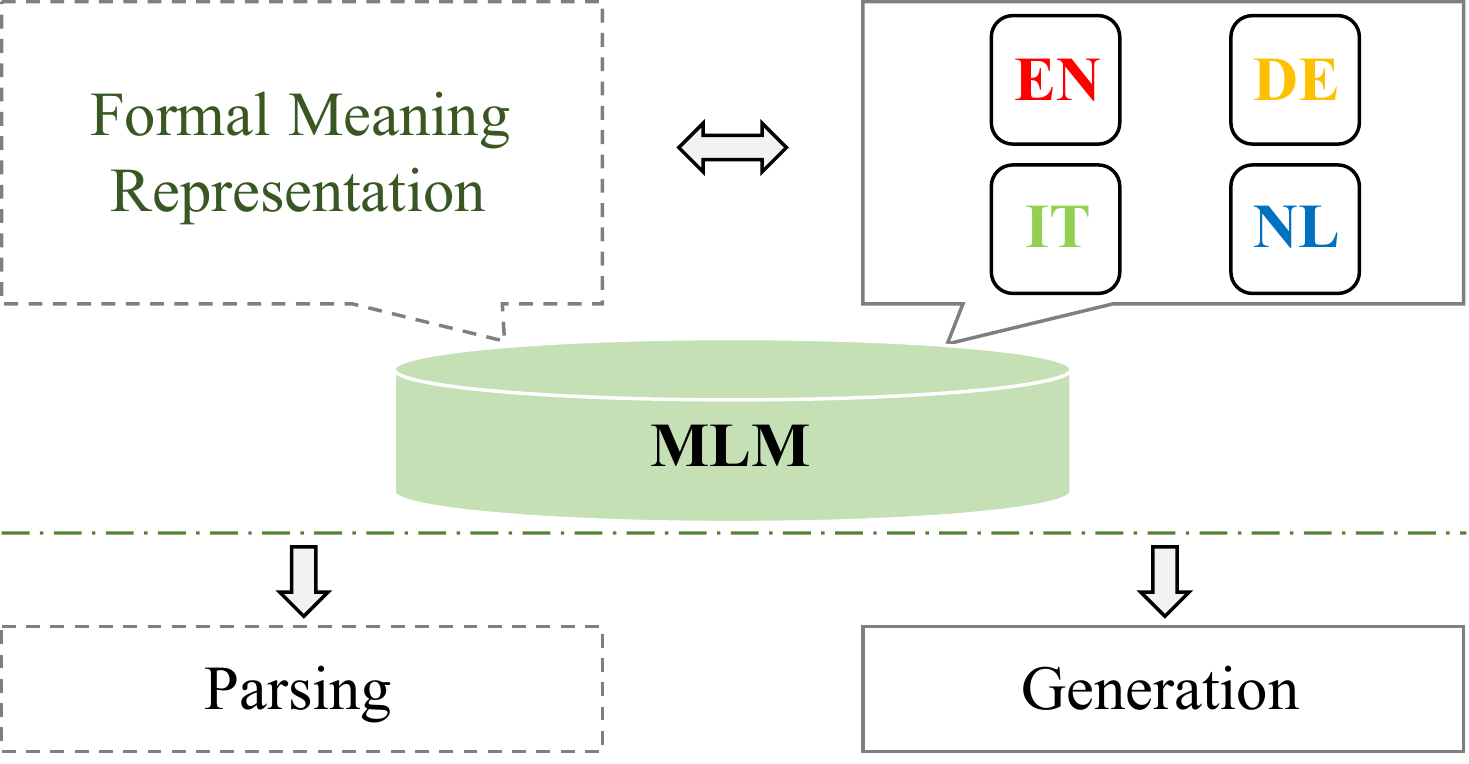}
\caption{Our \textbf{M}ultilingual (English:EN, German:DE, Italian:IT,  Dutch:NL) \textbf{L}anguage-\textbf{M}eaning framework (MLM) for parsing and generation.} 
\label{fig:idea}
\end{figure}

Our goal in this work is to leverage the principle of pre-trained models and explore the benefit of multilingual semantic parsing and generation of including  \textit{in the same model} meaning representations aside from natural language.  
This would make it possible not only to operate multilingually, thanks to representation neutrality, but also to leverage the bidirectionality of language-meaning alignment. 
Figure~\ref{fig:idea} illustrates our idea.

Semantic parsing and generation (in different languages) are clearly related, but traditionally they are studied and developed independently of one another, usually focusing on a single language (often English). This results in having to train separate models from scratch for each task and language, and progress has been hampered by data scarcity. This is especially true for languages other than English, where data scarcity is even more severe.

Our proposal to incorporate meaning representations in PLMs and to concurrently embrace a multilingual approach breaks with this tradition yielding a twofold advantage.
First, multilingual PLMs enable different languages to be represented in one universal space making it possible to benefit from cross-lingual knowledge transfer in semantic parsing and generation. Second, joining the formal and natural language representations in training makes it possible to leverage one and the same model for parsing and generation. For this approach to work, we need a meaning representation framework where (i) the formalism is language-neutral, (ii) there is aligned data both in terms of meaning-language(s), but also multilingually across different languages, and (iii) there is enough expressivity to cover for a wide range of language phenomena.


Discourse Representation Structure (DRS), which satisfies our requirements well, is the formal meaning representation proposed in 
Discourse Representation Theory (DRT,~\citealt{Kamp1981-KAMATO-2,asher:drt,Kamp1993,kadmon:drt,genabith:drt,sep:drt}). It covers a large variety of linguistic phenomena, including anaphors, presuppositions, temporal expressions and multi-sentence discourses and captures the semantics of negation, modals and quantification. Furthermore, DRS provides a language-neutral meaning representation: the same meaning representation associated with text that can be expressed in various languages.
While Abstract Meaning Representations (AMR,~\citealt{banarescu-etal-2013-abstract}) have been proposed for this task, we believe DRS is more suitable because of its multi-lingual representation capability  (all predicates are interpreted), its expressive power (proper treatment of negation and universal quantification), and the comparable annotated data available for multiple languages.  



As a first step, we consider DRS as an additional abstract language that will complement the natural languages in our pre-trained model.
We take the multilingual PLM mBART~\citep{liu-etal-2020-mbart-denoising} and further pre-train it with all of our language data, thus both the four natural languages we use as well as the language neutral meaning representations, so that the DRSs and texts are learnt in the same semantic space. As a second step, we introduce a supervised denoising training that exploits more explicitly the relationship between DRS and each corresponding text as well as between the parallel texts in the different languages; we do this combined with denoising training to reduce the gap between the pre-training and fine-tuning objectives. 
At this point, we have at our disposal a single multilingual language-meaning model which can then be fine-tuned for either parsing (text-to-DRS) or generation (DRS-to-text), in a monolingual or multilingual fashion.



Overall,  our main  contributions include: 
    (i) A novel task of multilingual DRS-to-text generation, and a framework for a mixed  language-meaning modelling in a multilingual setting, serving both parsing and generation.
    (ii) A pre-training strategy, with self-supervised training followed by supervised training, to reduce the gap between pre-training and fine-tuning; we also employ multilingual transfer techniques to boost performance in languages other than English exploiting language neutrality in DRSs.
    (iii) Extensive experiments for both parsing and generation across different languages, including both automatic and human evaluation to understand how multilingual models perform.\footnote{Code and models are available at \url{https://github.com/wangchunliu/DRS-pretrained-LMM}.}


\section{Background and Related Work}

This work employs intensive multilingual pre-training techniques for language-meaning modelling for both parsing and generation. In this section, we briefly introduce the concept of DRS, which serves as our meaning representation tool, and relevant background and related work.

\paragraph{Discourse Representation Structures}

\begin{figure}[!t]
\centering
\includegraphics[scale=.68]{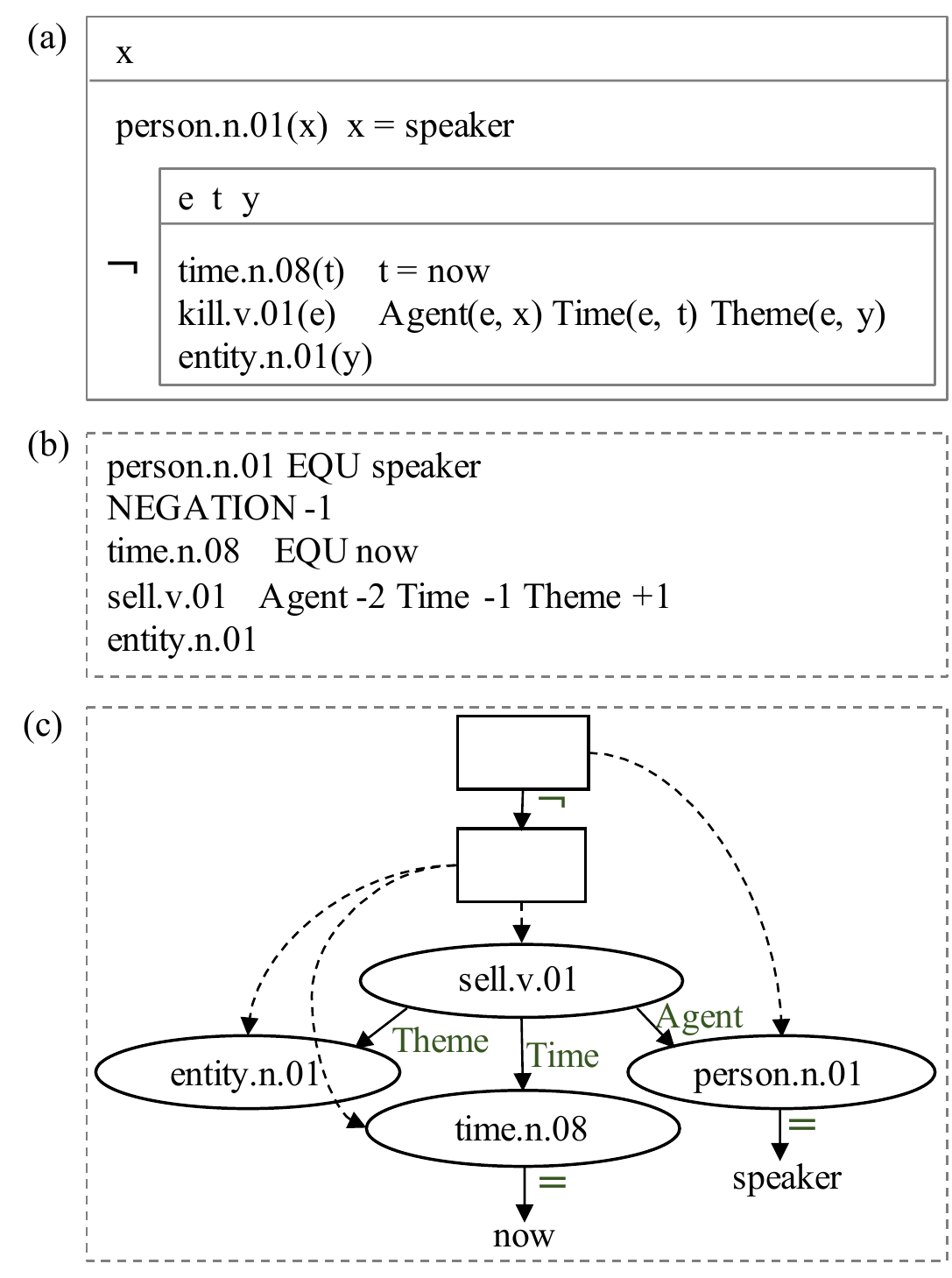}
\caption{Example of different formats of DRS for the English sentence "I'm not selling anything": the box format (a), the variable-free sequence notation (b), and as directed acyclic graph (c).} 
\label{fig:drs}
\vspace*{-10pt}
\end{figure}

The Parallel Meaning Bank (PMB,~\citealt{lasha-pmb})  
provides a large corpus of sentences annotated with DRSs in different formats for three different degrees of annotation quality: 
gold (completely checked manually), silver (partially checked manually) and bronze (uncorrected).\footnote{See \url{https://pmb.let.rug.nl/data.php}.}
The box-format of DRS extensively used in Discourse Representation Theory may be convenient for human readability, but it is not suitable for modelling. We thus use the Discourse Representation Graph (DRG) format provided by the PMB and its equivalent  variable-free sequential notation (Figure~\ref{fig:drs}).
There are three types of nodes in a DRG, a  directed acyclic graph: 
conceptual entities (represented by WordNet \cite{wordnet} synsets), 
constants (names, quantities, and the discourse deictics \texttt{speaker}, \texttt{hearer}, and \texttt{now}),
contexts (defining scope as a box in DRT, represented graphically as a box).
Edges between entity nodes denote thematic roles (Agent, Theme, Patient, Experiencer, Stimulus, Time, etc.) and 
comparison operators (=, $\neq$, $\prec$, $\leq$, $\sim$, and so on); edges between context nodes are discourse relations including negation (Figure~\ref{fig:drs}).

Even though the PMB resorts to the English version of Wordnet \cite{wordnet}, we consider a synset as an interlingual way of representing a concept, being a compound of a lemma, part of speech (noun, verb, adjective, adverb) and sense number. This means that DRSs for languages other than English also employ the synsets of the English WordNet as a sort of interlingua.
Only names in a DRS are language-specific --- for instance, the city of London would be represented in an Italian DRS as 
\texttt{city.n.01 Name "Londra"}.

The sequence notation for DRGs is based on a variable-free representation of DRS \citep{bos2021variable}.
In this notation a DRS is just a sequence of conceptual entities, roles with hooks (indices) or anchors, and discourse relations.
Each entity is followed by the roles it introduces. Each thematic role or comparison operator either hooks to another entity via a negative or positive index ($-1$ relates to the previous entity in the sequence, $-2$ to the one before that, $+1$ to the next one, and so on). Discourse relations (e.g., NEGATION, NARRATION, ELABORATION) in the sequence notation introduce new contexts (see Figure~\ref{fig:drs}).   We make heavily use of this sequential notation because of the many advantages it offers. For example, compared with the box-format DRS, it can be easily converted into a graph structure without the complicated conversion process introduced in previous work \citep{fancellu-2019-semantic, fu-etal-2020-drts}. 
Compared with the clause-format DRS~\citep{van-2018-exploring}, it omits the use of variables and is therefore simpler.  It can also be used directly to train a sequence-to-sequence (seq2seq) neural model.

\paragraph{Text-to-DRS Parsing}

In the traditional efforts for DRS parsing, it can be roughly divided into two categories, namely rule-based and neural network-based methods. Regarding rule-based methods, Boxer~\citep{bos08-boxer} is a classic system based on rules and statistical methods. Recently,~\citet{wessel2022} propose a multilingual DRS parser leveraging existing off-the-shelf Universal Dependency parsers, it can achieve similar or even better performances than BERT-based models. Indeed, neural models have become the most popular methods in this field and usually achieve the best performance~\citep{van-2018-exploring, liu-2019-discourse-representation, evang-2019-transition, van-2019-linguistic, van-2020-character, wang-etal-2021-input}. 
In addition to the seq2seq models above, there are two lines focusing on tree-based approaches \citep{liu-2018-discourse, liu-etal-2019-discourse} and graph-based approaches \citep{fancellu-2019-semantic, fu-etal-2020-drts}, where \citet{fancellu-2019-semantic} is the first attempt at multilingual DRS parsing. 

Most of the above works train neural models from scratch, and some make use of PLMs, but the models do not contain meaning representations explicitly during pre-training. Therefore, we aim to leverage the principle of pre-trained models and incorporate both meaning representations and natural language into one model. This, hopefully, can enable different languages to be represented explicitly in one universal space through pre-training, and result in one model for parsing and generation.




\paragraph{DRS-to-Text Generation}

Compared to DRS parsing, DRS-to-text generation has only recently drawn interest from NLP practitioners \citep{basile-bos-2011-towards,narayan-gardent-2014-hybrid, basile-2015-generation}. Similar to DRS parsing, prior work on the generation task can be classified into rule-based methods \citep{basile-bos-2011-towards} and neural network-based methods \citep{liu-2021-generation, wang-2021-evaluating}. 
All these works focus on English only. Here, we take the first step towards a multilingual generation task and provide a corresponding benchmark, leveraging the representation neutrality in DRS and the bidirectionality of language-meaning alignment in different languages. 


\paragraph{Multilingual Pre-Training} 


In recent years, multilingual PLMs have brought NLP to a new era~\citep{liu-etal-2020-mbart-denoising, xipeng2020, xue-etal-2021-mt5}.
They are pre-trained on large-scale unlabeled data in a self-supervised way, which enable different languages to be represented in one semantic space. Therefore, 
models fine-tuned on high-resource languages 
can thus transfer knowledge to other lower-resource languages for various tasks, such as Natural Language Inference~\citep{conneau-etal-2018-xnli}, Question Answering~\citep{clark-etal-2020-tydi}, Machine Translation~\citep{liu-etal-2020-mbart-denoising}, and formality transfer~\citep{lai-etal-2022-multilingual}. 



Generally, PLMs are pre-trained in a self-supervised manner, which enforces models to reconstruct corrupted text based on denoising objectives~\citep{liu-etal-2020-mbart-denoising}. However, recent work shows that self-supervised pre-training may introduce noisy information that affects the performance of downstream tasks~\citep{feng2022rethinking, tang2022mvp}. Moreover, it has been shown that supervised pre-training can achieve superior performance compared to the self-supervised approaches \citep{alexis2019, tang2022mvp}. In terms of computational semantics, ~\citet{bai-etal-2022-graph} propose a monolingual framework based on AMR, where the pre-training and fine-tuning share the same data format to facilitate knowledge transfer between them. 
Inspired by these works, we model meaning representations and natural language jointly leveraging the principle of PLMs in a multilingual fashion, and propose a pre-training strategy to make the pre-training objectives close to target downstream tasks by exploiting the relationship between DRS and its corresponding texts in different languages.

\begin{figure}[!t]
\centering
\includegraphics[scale=.5]{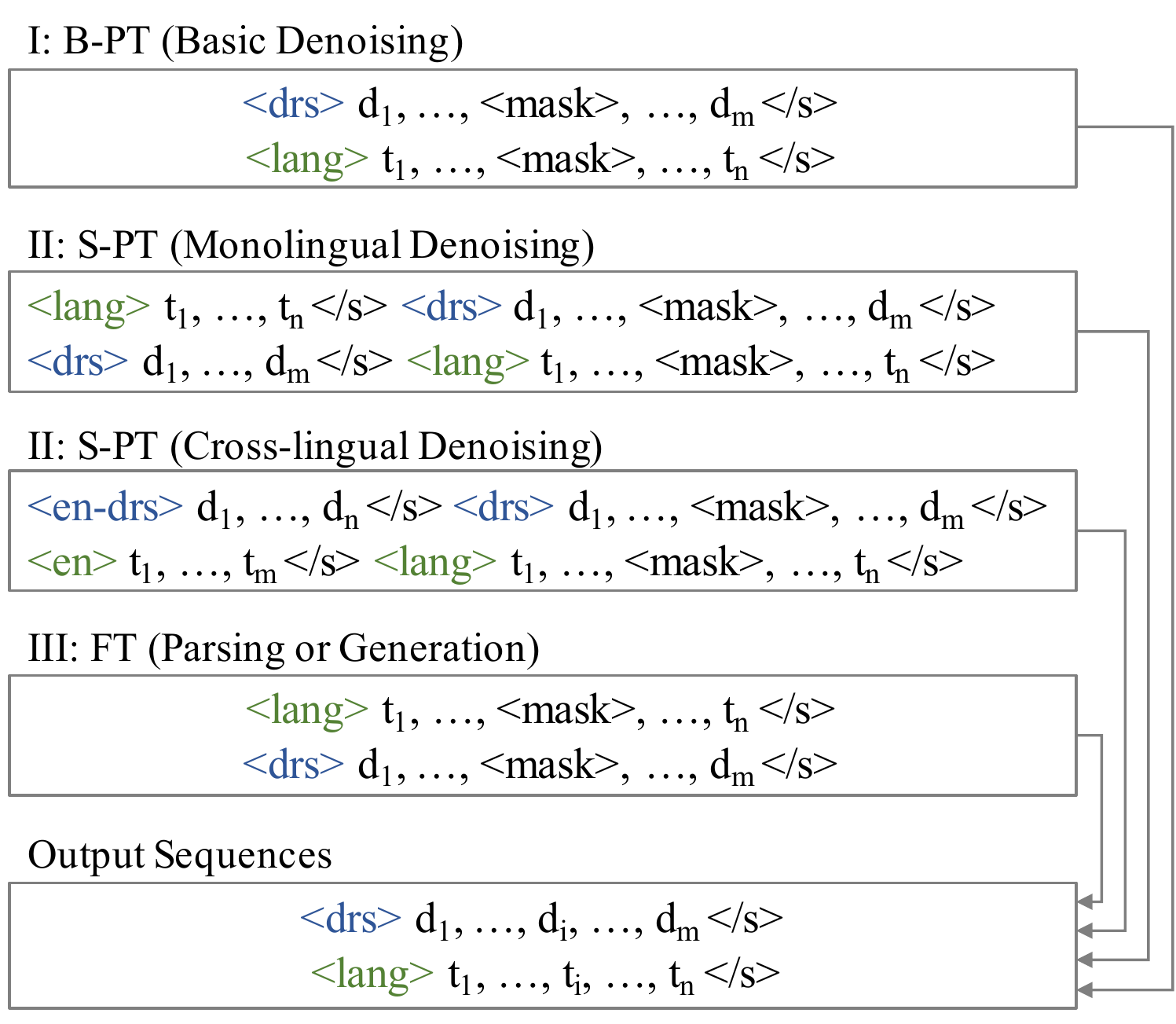}
\caption{Pre-training and fine-tuning strategies for the language-meaning model. In the B-PT stage, the model is trained with basic denoising. The S-PT stage contains both monolingual and cross-lingual objectives.}
\label{fig:overview} 
\vspace*{-5mm}
\end{figure}

\section{Method}

We use mBART as our backbone to jointly model natural language and meaning representation in a multilingual manner, thereby enabling the DRS representations and the texts to be learnt in the same semantic space. This one model is then fine-tuned for parsing and generation.

\subsection{mBART}

mBART is a pre-trained denoising seq2seq model based on the Transformer architecture~\citep{vaswani2017attention}, derived from the monolingual model BART~\citep{lewis-etal-2020-bart}. 
It is pre-trained to reconstruct the original text from a corrupted version (e.g.\ token masking). 
The model then takes the original sequence as input and maps it into the target sequence during fine-tuning and inference on downstream tasks. 
The novelty of our approach relies on the fact that the sequential DRS format allows for both text-to-DRS parsing and DRS-to-text generation to be performed in a seq2seq way (see Figure~\ref{fig:evaluation}). 
For more efficient training, we filter out the unused tokens from mBART's vocabulary after tokenizing the training corpora (including texts and DRSs), which results in a shared vocabulary of 39,981 tokens. Besides, we add a special token \texttt{<drs>} as a prefix for DRSs, which is used to distinguish DRSs from natural languages and guide models to produce DRSs as outputs of parsing.

\subsection{Multilingual Language-Meaning Models}

We introduce a pre-training strategy to model natural language and meaning representation on top of mBART, including (i) basic denoising training and (ii) supervised denoising training.

\paragraph{Basic Denoising Training} 

Since the meaning representations are not included in vanilla mBART, we perform a further pre-training to incorporate DRSs into the model and learn the universal representation.  Specifically, we combine all the training data of multiple languages: $\bm{D} = \{D_{1}, ..., D_{n}\}$ where each $D_{i}$ is a collection of data in a language. Language code \texttt{<lang>} and DRS code \texttt{<drs>} are used as prefixes for text and DRS sequences, respectively, to differentiate them from each other. As shown in Figure~\ref{fig:overview} (I: B-PT block), we follow~\citet{liu-etal-2020-mbart-denoising} to conduct a denoising training, which aims to reconstruct the original sequence from a version corrupted with a noise function. Formally, this denoising training can be formulated as:
\begin{equation}\label{eq:denoising}
    L_{\theta} = -\sum \log (T\mid g(T); \theta)
\end{equation}

\noindent where $\theta$ are the parameters of mBART and $g$ is the noise function that masks 35\% of tokens in each sequence at random.

\paragraph{Supervised Denoising Training} 

Although the basic denoising training makes the model learn the representations for text and DRS in a universal space, during this process  
the specific relationship between a given DRS and its corresponding texts is not learnt.
There is thus a gap between the denoising pre-training and the fine-tuning for the text-to-DRS and DRS-to-text downstream tasks.


To bridge this gap, we perform a supervised denoising training using all parallel language-meaning pairs. This enables our model to learn the transformation connection between text and DRS after the first step of basic denoising training. As shown in Figure~\ref{fig:overview} (II: S-PT block), we concatenate the text sequences with the corresponding corrupted DRS sequences and conduct denoising training to reconstruct the original DRS in the text-to-DRS direction, and vice versa. Inspired by~\citet{wang-etal-2022-training}, who show that retrieving and concatenating training instances relevant to the input can lead to significant gains on language generation tasks, we also perform an English-centric cross-lingual denoising training: English text (or DRS) sequences are concatenated with their corresponding corrupted non-English text (or DRS) sequences and then used for supervised denoising training (and vice versa).

\subsection{Parsing and Generation}

After denoising pre-training, the single model we have obtained can be fine-tuned with DRS-text pairs for the downstream DRS parsing and DRS-to-text generation tasks. As shown in Figure~\ref{fig:overview} (III: FT block), 
given a sequence $\bm{d} = \{d_{1}, \cdots, d_{n}\}$ of DRS and its corresponding text sequence $\bm{t} = \{t_{1}, \cdots, t_{m}\}$, taking DRS-to-text generation as an example, its seq2seq training can be formulated as follows:
\begin{equation}
p_{\theta}(\bm{t} | \bm{d}) 
= \prod_{i=1}^{m} p_{\theta}(t_i | t_{1,...,i-1}; \bm{d})
\end{equation}

\noindent Similar to previous work \citep{van-noord-etal-2020-character, wang-etal-2021-input}, we first train the model on gold + non-gold data, and then on gold + silver data. 

In the first step (F-FT), we use the multilingual DRS-text pairs from dataset $\bm{D}$ since the same meaning representation can be expressed in various languages. We expect that this process can allow the model to further benefit from knowledge transfer across different languages. After that, the model can be finally fine-tuned on silver and gold data in either a multilingual or monolingual manner (S-FT).




\section{Experiments}

For all experiments we use PMB release 4.0.0, which contains texts in English, German, Dutch and Italian for three levels of annotation (gold, silver and bronze).
Table~\ref{table:number_data} shows the statistics for the various languages, where each counted instance is a sentence and its corresponding DRS. (A small portion of DRSs that cannot be converted to DRGs were removed from the data set.)

\begin{table}[!t]
\centering
\setlength{\tabcolsep}{6pt}
\resizebox{\columnwidth}{!}{
\begin{tabular}{lrrrrr}
\toprule
 \textbf{Data type} & \multicolumn{3}{c}{\textbf{Gold}}  &\textbf{Silver}  & \textbf{Bronze}   \\
\hline
\textbf{Lang} & \textbf{Train} & \textbf{Dev} &\textbf{Test} & \textbf{Train}  & \textbf{Train}   \\
\midrule
English & 8,407 & 1,147 & 1,042 & 119,002 & 148,164\\
German  & 1,730 & 552   & 545   & 5,986   & 140,654\\
Italian & 682   & 540   & 459   & 3,995   & 98,382 \\
Dutch   & 535   & 435   & 490   & 1,363   & 26,433 \\
\bottomrule
\end{tabular}}
\caption{Documents statistics for PMB release 4.0.0.}
\label{table:number_data}
\end{table}

\begin{table}[t]
\centering
\resizebox{\columnwidth}{!}{
\begin{tabular}{lcccc}
\toprule
Hyper-Parameter & B-PT & S-PT & F-FT & S-FT\\
\hline
Batch size & 32 & 32 & 32 & 32\\
Update Steps & 8 & 8 & 8 & 1\\
Max learning rate & 1e-4 & 1e-5 & 5e-5 & 1e-5\\
Min learning rate & 1e-5 & 1e-5 & 1e-5 & 1e-5\\
Warmup updates & 3,000 & 0 & 3,000 & 0\\
Max decay steps & 30,000 & 0 & 30,000 & 0\\
\bottomrule
\end{tabular}}
\caption{Detailed hyper-parameters in our experiments.}
\label{table:hyper-parameters}
\end{table}

\subsection{Training Details}
\label{app:details}

Table~\ref{table:hyper-parameters} reports the detailed hyper-parameters in our experiments.
All experiments are implemented atop the Transformers library~\citep{wolf-etal-2020-transformers}. We use mBART-50~\citep{tang2020multilingual} as our base model, and train our models with batch size 32, accumulating gradients over 8 update steps in all training except for monolingual fine-tuning which is 1. We use Adam optimiser~\citep{kingma2017adam} with a polynomial learning rate decay. 
Additionally, we apply early stopping (patience 5) if validation performance does not improve. 
Due to the small size of the Dutch dataset, we upsample them by replication obtaining training sets of 100,000 DRS-text pairs in both pre-training and multilingual fine-tuning.

\subsection{Model Settings}

To show the effects of each training stage in our framework, we conduct extensive experiments with different settings, yielding five different models.
\textbf{M1} (FT mBART): fine-tuning vanilla mBART with monolingual data for each task; \textbf{M2} (M1 + B-PT): including basic denoising pre-training before monolingual fine-tuning; \textbf{M3} (M2 + S-PT): including supervised pre-training before monolingual fine-tuning, after basic pre-training; 
\textbf{M4} (M3 + F-FT): based on M3, and includes first multilingual fine-tuning (F-FT) before monolingual fine-tuning (S-FT); 
\textbf{M5} (monolithic model): based on M4, but using multilingual fine-tuning for S-FT and combining parsing and generation.

For comparison with our models, we also include two parsing systems from \citet{wessel2022} which use the same DRS data format as we do:
(i) UD-Boxer is a rule-based DRS parser based on Universal Dependencies;
 (ii) Neural Boxer is a seq2seq semantic parser based on Bi-LSTM with mBERT embeddings.

\subsection{Automatic Evaluation}

\begin{table*}[!t]
\centering
\setlength{\tabcolsep}{10pt}
\resizebox{\textwidth}{!}{
\begin{tabular}{l|cc|cc|cc|cc}
\toprule
\makecell[c]{\multirow{2}{*}{\textbf{Model}}} &  \multicolumn{2}{c|}{\textbf{EN}} & \multicolumn{2}{c|}{\textbf{DE}} & \multicolumn{2}{c|}{\textbf{IT}} &  \multicolumn{2}{c}{\textbf{NL}}\\
\cline{2-9}
 &F1 & ERR   &F1 &  ERR   &F1 &  ERR  &F1 & ERR\\
\midrule
\textbf{M1}: FT mBART    & 94.6 & 0.3 & 90.3 & 0.4 & 90.7 & 0.9 & 86.9 & 1.2\\
\textbf{M2}: M1 + B-PT   & \textbf{94.7} & 0.3 & 90.6 & 0.8 & 90.7 & 1.0 & 85.9 & 2.4 \\
\textbf{M3}: M2 + S-PT   & 94.6 & 0.3 & 91.3 & 0.9 & 90.9 & 0.7 & 88.2 & 1.6 \\
\textbf{M4}: M3 + F-FT   & 94.5 & 0.4 & \textbf{92.0} & 0.8 & 92.8 & 0.2 & 92.1 & 0.2\\
\textbf{M5}: monolithic model  & 94.0 & 0.2 & \textbf{92.0} & 0.4 & \textbf{93.1} & 0.2 & \textbf{92.6} & 0.6\\
\hline
UD-Boxer~\citep{wessel2022}             & 81.8 & \textbf{0.0} & 77.5 & \textbf{0.0} & 79.1 & \textbf{0.0} & 75.8 & \textbf{0.0} \\
Neural Boxer~\citep{wessel2022}         & 92.5 & 2.3 & 74.7 & 0.5 & 75.4 & \textbf{0.0} & 71.6 & 1.0 \\
\bottomrule
\end{tabular}}
\caption{Evaluation results for text-to-DRS parsing on the test set of the four languages in the PMB 4.0.0. 
Notes: (i) ERR is the ill-formed rate (\%) of generated DRSs that can not be transformed into a graph structure; (ii) bold numbers indicate best systems 
for each language.}
\label{table:auto-parsing}
\end{table*}

\begin{table*}[!t]
\centering
\setlength{\tabcolsep}{6pt}
\resizebox{\linewidth}{!}{
\begin{tabular}{l|ccc|ccc|ccc|ccc}
\toprule
\makecell[c]{\multirow{2}{*}{\textbf{Model}}}&  \multicolumn{3}{c|}{\textbf{EN}} & \multicolumn{3}{c|}{\textbf{DE}} & \multicolumn{3}{c|}{\textbf{IT}} &  \multicolumn{3}{c}{\textbf{NL}}\\
\cline{2-13}
 & B & M & C & B & M & C & B & M & C & B & M & C \\
\midrule
\textbf{M1}: FT mBART   & \textbf{74.5} & 54.7 & 102.8 & 45.1 & 35.1 & 54.3 & 44.3 & 34.4 & 58.2 & 34.9 & 29.5 & 31.3\\
\textbf{M2}: M1 + B-PT  & 73.2 & 54.0 & 101.5 & 45.0 & 34.8 & 56.8 & 44.2 & 34.2 & 59.7 & 38.6 & 31.8 & 44.4\\
\textbf{M3}: M2 + S-PT  & 74.2 & 54.6 & 102.4 & 52.1 & 38.4 & 65.3 & 49.3 & 36.6 & 72.6 & 47.8 & 38.6 & 59.9\\
\textbf{M4}: M3 + F-FT  & \textbf{74.5} & 54.8 & 102.4 & \textbf{56.3} & \textbf{40.8} & \textbf{76.7} & \textbf{58.0} & \textbf{41.1} & \textbf{85.5} & \textbf{60.8} & \textbf{43.4} & \textbf{79.8}\\
\textbf{M5}: monolithic model & \textbf{74.5} & \textbf{55.0} & \textbf{102.9} & \textbf{56.3} & \textbf{40.8} & 75.9 & 56.3 & 40.1 & 85.0 & 59.0 & 42.6 & 76.7\\
\bottomrule
\end{tabular}}
\caption{Automatic evaluation results for DRS-to-text generation on the test sets of the four languages in the PMB 4.0.0 (B = BLEU; M = METEOR; C = COMET).}
\label{table:all-generation}
\end{table*}

\begin{figure}[!t]
\centering
\includegraphics[scale=.69]{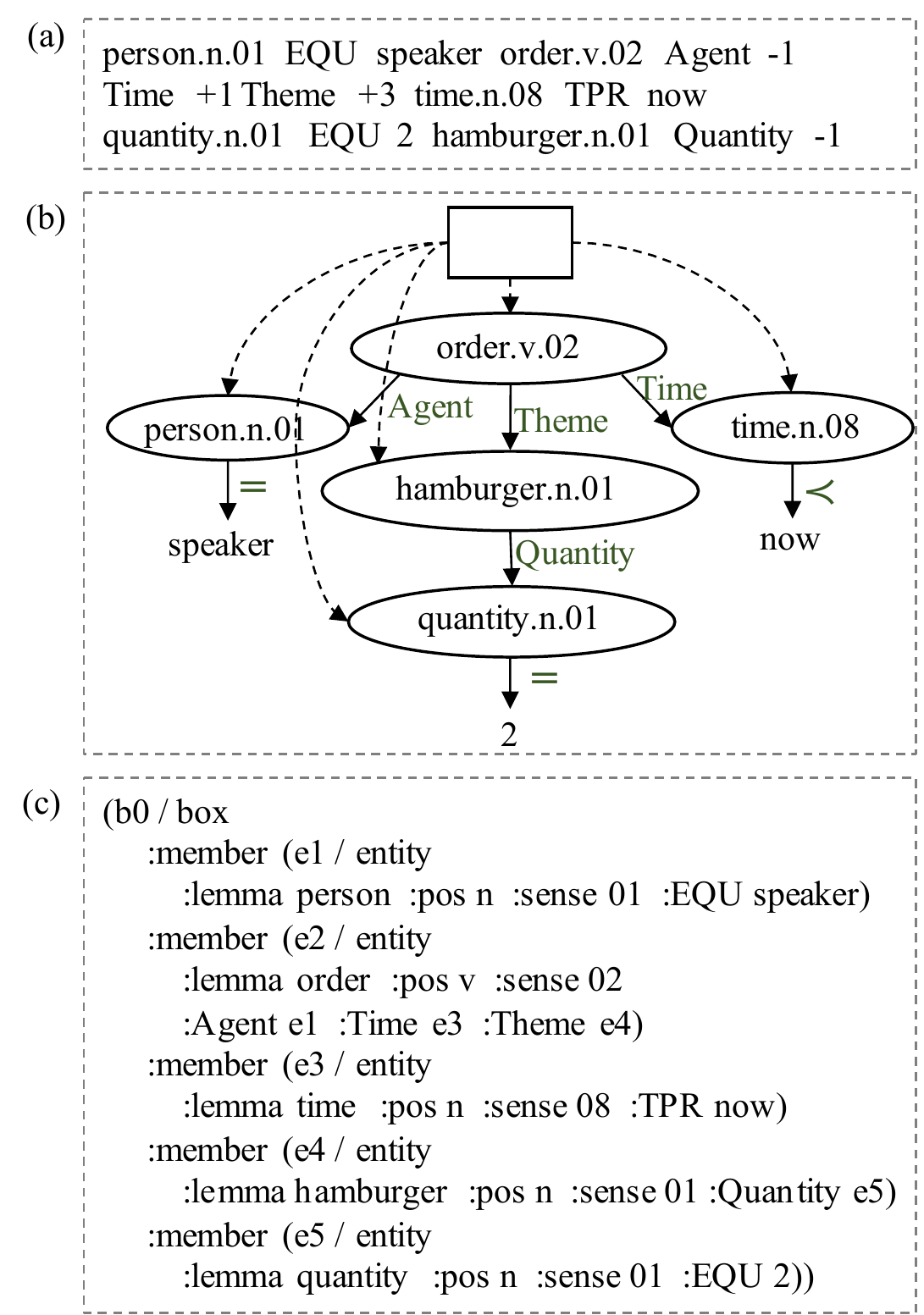}
\caption{Example of DRS parsing evaluation procedure for sentence \textit{I ordered two hamburgers}: linearized DRS data generated by parser (a), corresponding graphical DRG (b), and Penman format used for evaluation (c).}
\label{fig:evaluation}
\end{figure}

For \textbf{text-to-DRS parsing}, we follow recent work by \citet{wessel2022} to convert the linearized DRS into Penman format \citep{kasper-1989-flexible}, as shown in Figure~\ref{fig:evaluation}.
We then adopt Smatch, a standard evaluation tool used in AMR parsing, to compute overlap between system output and gold standard by calculating the F-score of matching triples \citep{cai-knight-2013-smatch}.

To assess \textbf{DRS-to-text generation}, we use three automatic metrics commonly used in text generation: 
$n$-gram-based BLEU \citep{2002bleu} and METEOR \citep{2007-meteor}, as well
as a neural-based COMET\footnote{We use model \texttt{wmt-large-da-estimator-1719}.}~\citep{rei-etal-2020-comet}.

\subsection{Automatic Evaluation Results}

Table~\ref{table:auto-parsing} reports the results of DRS parsing in different languages. For English, the performances of the different models are pretty close to each other, with M2 outperforming the others with basic pre-training and monolingual fine-tuning. The models show higher scores for English compared to the other three languages, most likely because the dataset contains a large amount of gold and silver DRS-text pairs in English, sufficient to fine-tune mBART for parsing without further pre-training.

When looking at the other three languages, we observe performance improvements with the use of different training strategies. Models pre-trained with the basic denoising task produce better results in German, the same F1-score in Italian, and lower results in Dutch, indicating a gap between pre-training and fine-tuning. 
This gap is bridged by our supervised pre-training strategy, models with the supervised pre-training (M3) yield steady improvements compared to M1 and M2.
For M4 fine-tuned with multilingual data, they can further benefit from cross-lingual knowledge transfer and achieve higher scores. It is interesting to see that our monolithic model~M5 performs best, thanks to the language-neutral meaning representation.

Compared to existing models UD-Boxer and Neural Boxer, all of our models, especially our main model (M5), achieve higher F1-scores across the board, showing significant improvements in four languages.
Our models perform worse than UD-Boxer in terms of ill-formedness rate, i.e., the proportion of generated DRSs which cannot be converted into a graph structure (and receive an F-score of 0). 
It is perhaps not surprising that rule-based parsers outperform neural-based parsers in generating well-formed DRS: the UD-Boxer parser is based on Universal Dependency and adds manual transformation rules to finally get the linearized data from the graph structure, and the evaluation process is equivalent to a reverse transformation process. It is worth noting that most of these errors can be corrected by post-processing (see \S\ref{sub:case}).
We also observe that our models have lower ERR rates than Neural-Boxer, except for Italian.
The possible reason for this is that the multilingual training may introduce some noise.

For the generation task, we observe similar trends to parsing, as shown in Table~\ref{table:all-generation}. 
Concretely, Our proposed supervised denoising pre-training and multilingual fine-tuning strategies substantially boost the performances, especially non-English languages. Model~M4 has the highest scores in all evaluation metrics across the three languages, the observation that differs slightly from that for the parsing task. 
We believe the reason is that the output tokens of the generation task are language-particular rather than language-neutral compared to the parsing task. Therefore, for the generation task, the results of fine-tuning with monolingual data are better than those with multilingual data.

\section{Analysis}

\subsection{Correlation Analysis}

While human evaluation is seen as the most reliable assessment in language generation tasks, due to its costs and availability it can not be easily used during iterative development. We included human evaluation early on in our experiments to check the correlation of human judgement with automatic metrics, so that the latter could be  more safely used in the following stages of our experiments.\footnote{See Appendix~\ref{table:human-results} for the details on human evaluation.} 


\begin{table}[!t]
\centering
\setlength{\tabcolsep}{13pt}
\resizebox{\linewidth}{!}{
\begin{tabular}{cccc}
\toprule
\textbf{Lang} & \textbf{BLEU} & \textbf{METEOR} & \textbf{COMET} \\
\hline
  \textbf{EN} & -0.098 & -0.016 & \underline{0.775}\\
  \textbf{DE} & 0.275 & \underline{0.471} & \underline{0.687}\\
  \textbf{IT} & 0.122 & 0.241 & \underline{0.768}\\
  \textbf{NL} & 0.195 & \underline{0.386} & \underline{0.686}\\
\bottomrule
\end{tabular}}
\caption{Sentence-level correlations of automatic metrics (against human reference) and human judgments for semantics. Underlined scores indicate $p$ $<$ 0.01.}
\label{table:corr-results}
\end{table}


\begin{figure*}[t]
    \begin{minipage}[t]{0.65\linewidth}
    \centering
    \subfigure[DRS parsing.]{
      \includegraphics[scale=0.505]{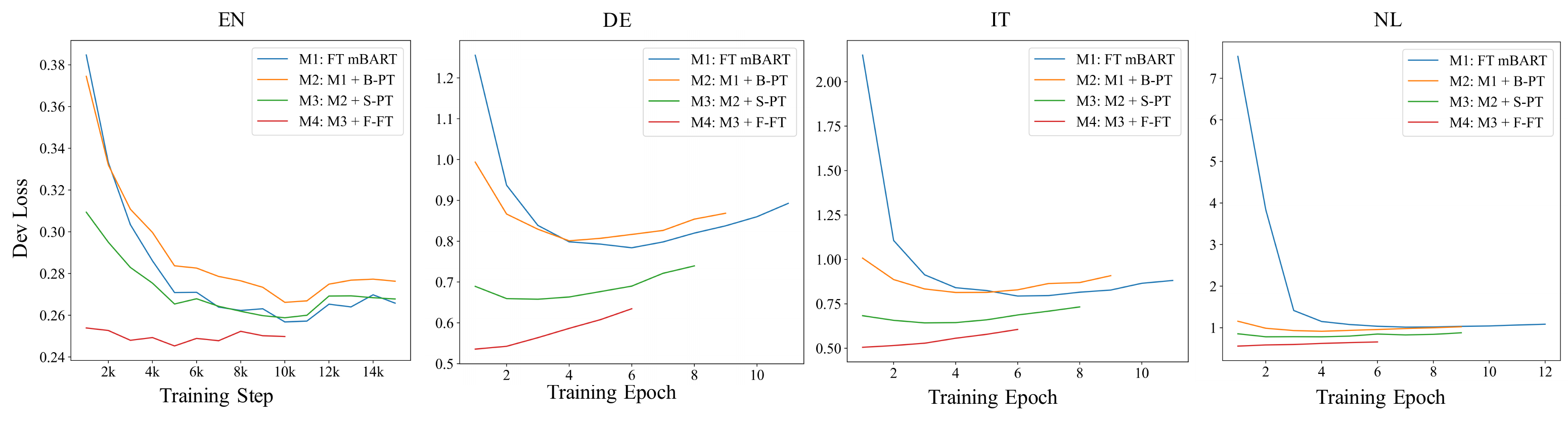}
      \label{fig:drs-loss}
    }
    \end{minipage}

    \vspace{1.0mm}
    \begin{minipage}[t]{0.65\linewidth}
    \centering
    \subfigure[DRS-to-text generation.]{
      \includegraphics[scale=0.505]{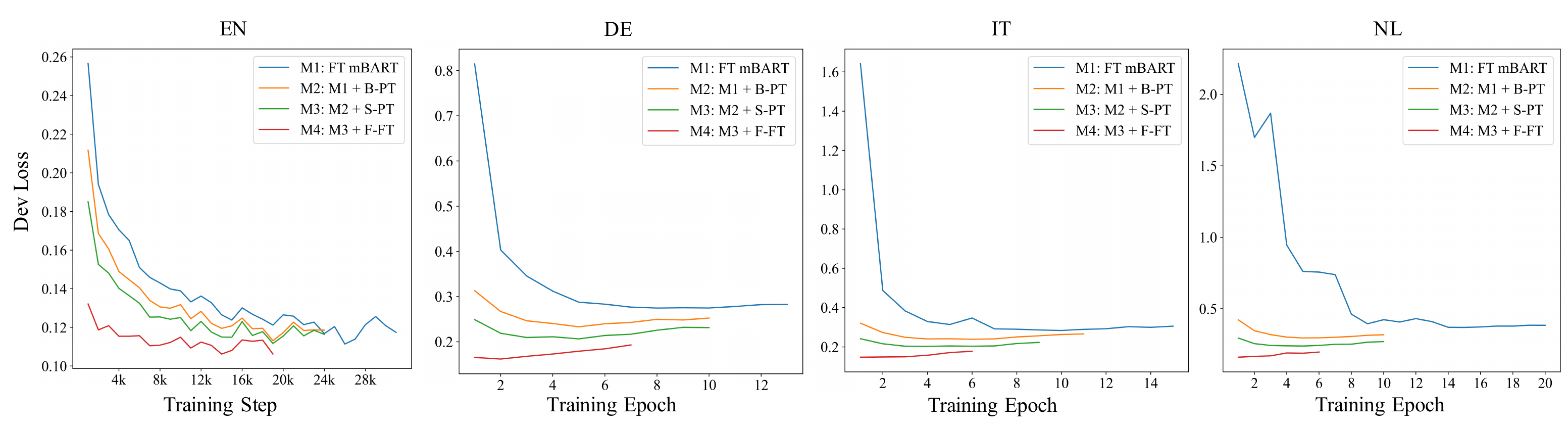}
      \label{fig:gen-loss}
    }
    \end{minipage}
    
    \caption{Loss curves of monolingual fine-tuning on the development sets.}
    \label{fig:dev-loss}
\end{figure*}

Table~\ref{table:corr-results} shows the sentence-level biserial correlations between automatic metrics and expert judgments in meaning preservation.\footnote{Since the biserial correlation coefficient is a statistic used to assess the degree of relationship between an artificially created dichotomous nominal scale and an interval scale, it is naturally applicable to our experiments as the generated text is rated by annotators with 0~or~1.} BLEU correlates particularly poorly with human judgments, even showing a negative correlation in English. METEOR also shows a negative correlation with human ratings in English, while it has higher scores than BLEU in non-English languages. Unsurprisingly, we see that COMET has high correlations with human judgements, which is consistent with previous work on other tasks~\citep{rei-etal-2020-comet, lai-etal-2022-human}. This observation, therefore, confirms that COMET can be a more reliable metric used for DRS-to-text generation and for comparisons between different models.


\begin{table*}[htb]
\centering
\setlength{\tabcolsep}{4pt}
\resizebox{\textwidth}{!}{
\begin{tabular}{llll}
\toprule
  \makecell[c]{\textbf{Type}} & \makecell[c]{\textbf{Subtype}}& \makecell[c]{\textbf{Output Meaning}}  & \makecell[c]{\textbf{Gold Meaning}}  \\
\midrule
\multirow{3}{*}{Ill-formed} 
 & \multirow{2}{*}{Extra Space} 
 &  geological\_formation.n.01 Name \textcolor{red}{" Himalayas"} &  geological\_formation.n.01 Name \textcolor{blue}{"Himalayas"} \\
& & \textcolor{red}{driving\_ licence.n.01} Owner speaker & \textcolor{blue}{driving\_licence.n.01} Owner speaker \\
&  Missing Space & person.n.01 Role \textcolor{red}{+1technician.n.01} & person.n.01 Role \textcolor{blue}{+1 engineer.n.01} \\
\hline
\multirow{5}{*}{Meaning} 
& \multirow{1}{*}{Wrong Concept}&  \textcolor{red}{overtreibe.v.01} Patient -1 Time +1  & \textcolor{blue}{exaggerate.v.01} Agent -1 Time +1  \\
& \multirow{1}{*}{Wrong Role}   &  blind.a.01 \textcolor{red}{Experiencer} -3 Time -2 &   blind.a.01 \textcolor{blue}{Theme} -3 Time -2 \\
& \multirow{1}{*}{Wrong Index}  &  female.n.02 Name "Maria" \textcolor{red}{EQU +1} EQU now  & female.n.02 Name "Maria" \\
& \multirow{1}{*}{Missing Token}  &  young.a.01 AttributeOf +1 person.n.01 & young.a.01 \textcolor{blue}{Value +} person.n.01 Attribute -1 \\
& \multirow{1}{*}{Extra Token} &  \textcolor{red}{more\_and\_more.a.01 Degree +1 more.r.01} & more\_and\_more.r.01 \\
\bottomrule
\end{tabular}}
\caption{Example outputs produced by our best model (M5) for the parsing task.}
\label{table:example-parsing}
\end{table*}

\begin{table*}[htb]
\centering
\setlength{\tabcolsep}{12pt}
\resizebox{\textwidth}{!}{
\begin{tabular}{llll}
\toprule
 \makecell[c]{\textbf{Reason}} & \makecell[c]{\textbf{Lang}} & \makecell[c]{\textbf{Generated Text}}  & \makecell[c]{\textbf{Gold text}} \\
\midrule
\multirow{1}{*}{Semantic} 
&  IT & Peter sta comprando un gatto \textcolor{red}{male}. & Peter sta comprando un gatto \textcolor{blue}{maschio}.\\
\hline
\multirow{1}{*}{Grammaticality} 
& NL &  Tom \textcolor{red}{foldt} zijn kleren. & 	Tom \textcolor{blue}{vouwt} zijn kleren op. \\
\hline
\multirow{1}{*}{Extra Material} &   EN &  My flight arrived \textcolor{red}{exactly} at 2:30 p.m. &  My flight arrived at 2:30 p.m.\\ 
\hline
\multirow{1}{*}{Missing Material} & EN &  The express arrives at 6:30. &  The express arrives at 6:30 \textcolor{blue}{p.m.} \\
\hline
\multirow{1}{*}{Word Choice} 
& NL & Charles de Gaulle \textcolor{red}{stierf} in 1970. & 	Charles de Gaulle \textcolor{blue}{overleed} in 1970. \\
\bottomrule
\end{tabular}}
\caption{Example outputs generated by our best model (M4) for the generation task.}
\label{table:example-generation}
\end{table*}

\subsection{Development Loss}
\label{app:dev-loss}

To better understand the training strategies and components in our proposed framework, we examine the loss curves for different monolingual fine-tuned models on the dev sets of different languages (Figure~\ref{fig:dev-loss}).

For DRS parsing, the convergence process of the original mBART (M1) is slow.  
After adding different training strategies, models have a significantly faster and better convergence process. Specifically, we observe that basic denoising pre-training makes the model learn the representation for DRSs and texts in the same semantic space, but there is still a gap between the basic denoising task and the downstream task. This gap is then eliminated by supervised pre-training as the loss of model~M3 is quite flat from the start and is lower than that of~M2. 
Lastly, we see that multilingual fine-tuning consistently helps the model, and it eventually converges fast and well. This suggests that this strategy helps models benefit from the cross-knowledge transfer. 

We observe similar trends for the DRS-to-text task, with a large fluctuation in the convergence process without pre-training. Overall, the loss curves for M4 are lower than other models.

\subsection{Manual Inspection}\label{sub:case}

In Table~\ref{table:example-parsing} we report example DRS outputs from our main model (M5) which differ from the gold standard. 
We summarize two types of ill-formed DRSs in linear format that cannot be converted to graph structures (and hence are not interpretable).
When more tokens are produced than expected, the result is often a sequence of tokens that does not correspond to the graph. For instance, spaces are included where they shouldn't be, or missing spaces cause subsequent tokens to be erroneously connected to each other. 

These syntactic error types occur in a very limited number of models and can be well resolved by post-processing. We focus on the types of errors that affect meaning.
We show five typical semantic error types at the bottom of Table~\ref{table:example-parsing} that affect the number of matching triples and may lead to a different meaning in gold data.
For example, out-of-vocabulary (OOV) words may cause the parser to generate concepts different from gold yielding incorrect meanings. Also, incorrect roles lead to changes in meaning and wrong indices produce different predicate-argument structures. 
Another problem is when the parser fails to generate a crucial token. 
In contrast, the parser may hallucinate tokens, which may be added in unexpected places.

In Table~\ref{table:example-generation}, we show some examples of DRS-to-text generation which differ from the gold output for various reasons.
The model might produce a word which does not convey the intended meaning. 
For example, in the IT example, the word ``male'' (EN: ``bad'') is generated in place of ``maschio'' (EN: ``male''), probably due to the homography of the words across the two languages, without any semantic correspondence. 
Another example of non-matching is grammatical agreement, which can be due to some underspecified phenomena in DRSs.
We also identify three more types: (1) the generated text has redundant information; (2) the generated data lacks some information; (3) the generated words are synonymous with those in the gold references. 
These types generally degrade automatic evaluation results but may not affect the performance of human evaluation. The generation of these cases is usually random and occurs in all models. Part of it is probably due to the OOV problem, and the rest is mainly related to the training data itself, because the same meaning representation can be paired with multiple expressions.

\section{Conclusion and Future Work}

Using DRS-based meaning representations as an additional language aside four different natural languages yields a novel multilingual pre-trained language-meaning model that can be fine-tuned for both semantic parsing and generation from formal meaning representations. By doing so, we achieve state-of-the-art performance on both tasks. Exploiting parallel data and DRS language neutrality is key to boost performance in lesser-resourced languages. 

We believe our approach can benefit from improvements in its current form, but also opens up to further research in language-meaning models.
Regarding future modelling directions, the contribution of graph structures should be further explored in future work. Specifically, it could be possibly to leverage the graph structure to mask tokens in a more meaningful and principled way, designing a denoising training using the rich linguistic phenomena expressed by DRSs.



\section*{Limitations}

A large part of the dataset that we used in our experiments are semantic annotations for relatively short sentences (as the examples show). So we don't know really know
how our multilingual pre-trained language-meaning modelling for DRS parsing and DRS-to-text generation will work on longer sentences.

In our experiments, we converted meaning representation in the sequence notation and modelled them with natural language texts in a seq2seq manner and masked tokens in the DRS sequence randomly. Perhaps a more natural way is to model DRSs as graph structures
and let training objectives directly utilize any structural information from DRS. A graph structure would also eliminate the explicit order of concepts that is present in the sequence notation.

Although we say that the DRSs are language-neutral, the concepts in the vocabulary are based on the English WordNet. As a result it might be the case that non-English words do not have a direct correspondence to an appropriate synset, but the number of such cases is likely very small. 
The only (trivial) language dependence in DRSs are literal occurrences of proper names in cases where they differ across languages (e.g., "London", "Londen", or "Londra"). One way to remedy this is to add alternative spellings to the meaning representation to make it completely interlingual.

\section*{Acknowledgments}

This work was funded by the NWO-VICI grant ``Lost in Translation---Found in Meaning'' (288-89-003) and the China Scholarship Council (CSC). We thank the anonymous reviewers of ACL 2023 for their insightful comments. 
We would also like to thank the Center for Information Technology of the University of Groningen for their support and for providing access to the Peregrine high performance computing cluster.


\bibliography{anthology,custom}

\begin{thebibliography}{50}
\expandafter\ifx\csname natexlab\endcsname\relax\def\natexlab#1{#1}\fi

\bibitem[{Abzianidze et~al.(2020)Abzianidze, Van~Noord, Wang, and
  Bos}]{lasha-pmb}
Lasha Abzianidze, Rik Van~Noord, Chunliu Wang, and Johan Bos. 2020.
\newblock The parallel meaning bank: A framework for semantically annotating
  multiple languages.
\newblock \emph{AMIM}, 25(2):45--60.

\bibitem[{Asher(1993)}]{asher:drt}
Nicholas Asher. 1993.
\newblock \emph{{Reference to Abstract Objects in Discourse}}.
\newblock Kluwer Academic Publishers.

\bibitem[{Bai et~al.(2022)Bai, Chen, and Zhang}]{bai-etal-2022-graph}
Xuefeng Bai, Yulong Chen, and Yue Zhang. 2022.
\newblock \href {https://doi.org/todo} {Graph pre-training for {AMR} parsing
  and generation}.
\newblock In \emph{Proceedings of the 60th Annual Meeting of the Association
  for Computational Linguistics (Volume 1: Long Papers)}, page todo, Online.
  Association for Computational Linguistics.

\bibitem[{Banarescu et~al.(2013)Banarescu, Bonial, Cai, Georgescu, Griffitt,
  Hermjakob, Knight, Koehn, Palmer, and
  Schneider}]{banarescu-etal-2013-abstract}
Laura Banarescu, Claire Bonial, Shu Cai, Madalina Georgescu, Kira Griffitt, Ulf
  Hermjakob, Kevin Knight, Philipp Koehn, Martha Palmer, and Nathan Schneider.
  2013.
\newblock \href {https://aclanthology.org/W13-2322} {{A}bstract {M}eaning
  {R}epresentation for sembanking}.
\newblock In \emph{Proceedings of the 7th Linguistic Annotation Workshop and
  Interoperability with Discourse}, pages 178--186, Sofia, Bulgaria.
  Association for Computational Linguistics.

\bibitem[{Basile(2015)}]{basile-2015-generation}
Valerio Basile. 2015.
\newblock \emph{From logic to language: Natural language generation from
  logical forms}.
\newblock Ph.D. thesis, University of Groningen.

\bibitem[{Basile and Bos(2011)}]{basile-bos-2011-towards}
Valerio Basile and Johan Bos. 2011.
\newblock \href {https://aclanthology.org/W11-2819} {Towards generating text
  from discourse representation structures}.
\newblock In \emph{Proceedings of the 13th {E}uropean Workshop on Natural
  Language Generation}, pages 145--150, Nancy, France. Association for
  Computational Linguistics.

\bibitem[{Bos(2008)}]{bos08-boxer}
Johan Bos. 2008.
\newblock \href {https://aclanthology.org/W08-2222} {Wide-coverage semantic
  analysis with {B}oxer}.
\newblock In \emph{Semantics in Text Processing. {STEP} 2008 Conference
  Proceedings}, pages 277--286. College Publications.

\bibitem[{Bos(2021)}]{bos2021variable}
Johan Bos. 2021.
\newblock Variable-free discourse representation structures.
\newblock \emph{Semantics Archive}.

\bibitem[{Cai and Knight(2013)}]{cai-knight-2013-smatch}
Shu Cai and Kevin Knight. 2013.
\newblock \href {https://aclanthology.org/P13-2131} {{S}match: an evaluation
  metric for semantic feature structures}.
\newblock In \emph{Proceedings of the 51st Annual Meeting of the Association
  for Computational Linguistics (Volume 2: Short Papers)}, pages 748--752,
  Sofia, Bulgaria. Association for Computational Linguistics.

\bibitem[{Clark et~al.(2020)Clark, Choi, Collins, Garrette, Kwiatkowski,
  Nikolaev, and Palomaki}]{clark-etal-2020-tydi}
Jonathan~H. Clark, Eunsol Choi, Michael Collins, Dan Garrette, Tom Kwiatkowski,
  Vitaly Nikolaev, and Jennimaria Palomaki. 2020.
\newblock \href {https://doi.org/10.1162/tacl_a_00317} {{T}y{D}i {QA}: A
  benchmark for information-seeking question answering in typologically diverse
  languages}.
\newblock \emph{Transactions of the Association for Computational Linguistics},
  8:454--470.

\bibitem[{Conneau and Lample(2019)}]{alexis2019}
Alexis Conneau and Guillaume Lample. 2019.
\newblock \href
  {https://proceedings.neurips.cc/paper/2019/file/c04c19c2c2474dbf5f7ac4372c5b9af1-Paper.pdf}
  {Cross-lingual language model pretraining}.
\newblock In \emph{Advances in Neural Information Processing Systems},
  volume~32. Curran Associates, Inc.

\bibitem[{Conneau et~al.(2018)Conneau, Rinott, Lample, Williams, Bowman,
  Schwenk, and Stoyanov}]{conneau-etal-2018-xnli}
Alexis Conneau, Ruty Rinott, Guillaume Lample, Adina Williams, Samuel Bowman,
  Holger Schwenk, and Veselin Stoyanov. 2018.
\newblock \href {https://doi.org/10.18653/v1/D18-1269} {{XNLI}: Evaluating
  cross-lingual sentence representations}.
\newblock In \emph{Proceedings of the 2018 Conference on Empirical Methods in
  Natural Language Processing}, pages 2475--2485, Brussels, Belgium.
  Association for Computational Linguistics.

\bibitem[{Evang(2019)}]{evang-2019-transition}
Kilian Evang. 2019.
\newblock \href {https://doi.org/10.18653/v1/W19-1202} {Transition-based {DRS}
  parsing using stack-{LSTM}s}.
\newblock In \emph{Proceedings of the {IWCS} Shared Task on Semantic Parsing},
  Gothenburg, Sweden. Association for Computational Linguistics.

\bibitem[{Fancellu et~al.(2019)Fancellu, Gilroy, Lopez, and
  Lapata}]{fancellu-2019-semantic}
Federico Fancellu, Sorcha Gilroy, Adam Lopez, and Mirella Lapata. 2019.
\newblock \href {https://doi.org/10.18653/v1/D19-1278} {Semantic graph parsing
  with recurrent neural network {DAG} grammars}.
\newblock In \emph{Proceedings of the 2019 Conference on Empirical Methods in
  Natural Language Processing and the 9th International Joint Conference on
  Natural Language Processing (EMNLP-IJCNLP)}, pages 2769--2778, Hong Kong,
  China. Association for Computational Linguistics.

\bibitem[{Fellbaum(1998)}]{wordnet}
Christiane Fellbaum. 1998.
\newblock Wordnet: An electronic lexical database.
\newblock \emph{The MIT Press, Cambridge, Ma., USA}.

\bibitem[{Feng et~al.(2022)Feng, Jiang, Tang, Jin, and
  Gao}]{feng2022rethinking}
Yutong Feng, Jianwen Jiang, Mingqian Tang, Rong Jin, and Yue Gao. 2022.
\newblock \href {https://openreview.net/forum?id=Jjcv9MTqhcq} {Rethinking
  supervised pre-training for better downstream transferring}.
\newblock In \emph{International Conference on Learning Representations}.

\bibitem[{Fu et~al.(2020)Fu, Zhang, Liu, and Zhang}]{fu-etal-2020-drts}
Qiankun Fu, Yue Zhang, Jiangming Liu, and Meishan Zhang. 2020.
\newblock \href {https://doi.org/10.18653/v1/2020.acl-main.609} {{DRTS} parsing
  with structure-aware encoding and decoding}.
\newblock In \emph{Proceedings of the 58th Annual Meeting of the Association
  for Computational Linguistics}, pages 6818--6828, Online. Association for
  Computational Linguistics.

\bibitem[{Geurts et~al.(2020)Geurts, Beaver, and Maier}]{sep:drt}
Bart Geurts, David~I. Beaver, and Emar Maier. 2020.
\newblock {Discourse Representation Theory}.
\newblock In Edward~N. Zalta, editor, \emph{The {Stanford} Encyclopedia of
  Philosophy}, spring 2020 edition. Metaphysics Research Lab, Stanford
  University.

\bibitem[{Kadmon(2001)}]{kadmon:drt}
Nirit Kadmon. 2001.
\newblock \emph{Formal Pragmatics}.
\newblock Blackwell.

\bibitem[{Kamp(1981)}]{Kamp1981-KAMATO-2}
H.~Kamp. 1981.
\newblock A theory of truth and semantic representation, 277-322, jag
  groenendijk, tmv janssen and mbj stokhof, eds.
\newblock In Jeroen Groenendijk, editor, \emph{Formal Methods in the Study of
  Language}. U of Amsterdam.

\bibitem[{Kamp and Reyle(1993)}]{Kamp1993}
Hans Kamp and U.~Reyle. 1993.
\newblock From discourse to logic: Introduction to model theoretic semantics of
  natural language, formal logic and discourse representation theory.
\newblock \emph{Language}, 71(4).

\bibitem[{Kamp et~al.(2011)Kamp, van Genabith, and Reyle}]{genabith:drt}
Hans Kamp, Josef van Genabith, and Uwe Reyle. 2011.
\newblock {Discourse Representation Theory}.
\newblock In Dov~M. Gabbay and Franz Guenthner, editors, \emph{Handbook of
  Philosophical Logic}, volume~15, pages 125--394. Elsevier, MIT.

\bibitem[{Kasper(1989)}]{kasper-1989-flexible}
Robert~T. Kasper. 1989.
\newblock \href {https://aclanthology.org/H89-1022} {A flexible interface for
  linking applications to {P}enman{'}s sentence generator}.
\newblock In \emph{Speech and Natural Language: Proceedings of a Workshop Held
  at Philadelphia, {P}ennsylvania, {F}ebruary 21-23, 1989}.

\bibitem[{Kingma and Ba(2015)}]{kingma2017adam}
Diederik~P. Kingma and Jimmy Ba. 2015.
\newblock \href {https://arxiv.org/pdf/1412.6980.pdf} {{Adam}: A method for
  stochastic optimization}.
\newblock In \emph{International Conference on Learning Representations}.

\bibitem[{Lai et~al.(2022{\natexlab{a}})Lai, Mao, Toral, and
  Nissim}]{lai-etal-2022-human}
Huiyuan Lai, Jiali Mao, Antonio Toral, and Malvina Nissim. 2022{\natexlab{a}}.
\newblock \href {https://doi.org/10.18653/v1/2022.humeval-1.9} {Human judgement
  as a compass to navigate automatic metrics for formality transfer}.
\newblock In \emph{Proceedings of the 2nd Workshop on Human Evaluation of NLP
  Systems (HumEval)}, pages 102--115, Dublin, Ireland. Association for
  Computational Linguistics.

\bibitem[{Lai et~al.(2022{\natexlab{b}})Lai, Toral, and
  Nissim}]{lai-etal-2022-multilingual}
Huiyuan Lai, Antonio Toral, and Malvina Nissim. 2022{\natexlab{b}}.
\newblock \href {https://aclanthology.org/2022.acl-short.29} {Multilingual
  pre-training with language and task adaptation for multilingual text style
  transfer}.
\newblock In \emph{Proceedings of the 60th Annual Meeting of the Association
  for Computational Linguistics (Volume 2: Short Papers)}, pages 262--271,
  Dublin, Ireland. Association for Computational Linguistics.

\bibitem[{Lavie and Agarwal(2007)}]{2007-meteor}
Alon Lavie and Abhaya Agarwal. 2007.
\newblock {METEOR}: An automatic metric for mt evaluation with high levels of
  correlation with human judgments.
\newblock In \emph{Proceedings of the Second Workshop on Statistical Machine
  Translation}, StatMT '07, page 228–231, USA. Association for Computational
  Linguistics.

\bibitem[{Lewis et~al.(2020)Lewis, Liu, Goyal, Ghazvininejad, Mohamed, Levy,
  Stoyanov, and Zettlemoyer}]{lewis-etal-2020-bart}
Mike Lewis, Yinhan Liu, Naman Goyal, Marjan Ghazvininejad, Abdelrahman Mohamed,
  Omer Levy, Veselin Stoyanov, and Luke Zettlemoyer. 2020.
\newblock \href {https://doi.org/10.18653/v1/2020.acl-main.703} {{BART}:
  Denoising sequence-to-sequence pre-training for natural language generation,
  translation, and comprehension}.
\newblock In \emph{Proceedings of the 58th Annual Meeting of the Association
  for Computational Linguistics}, pages 7871--7880, Online. Association for
  Computational Linguistics.

\bibitem[{Liu et~al.(2018)Liu, Cohen, and Lapata}]{liu-2018-discourse}
Jiangming Liu, Shay~B. Cohen, and Mirella Lapata. 2018.
\newblock \href {https://doi.org/10.18653/v1/P18-1040} {Discourse
  representation structure parsing}.
\newblock In \emph{Proceedings of the 56th Annual Meeting of the Association
  for Computational Linguistics (Volume 1: Long Papers)}, pages 429--439,
  Melbourne, Australia. Association for Computational Linguistics.

\bibitem[{Liu et~al.(2019{\natexlab{a}})Liu, Cohen, and
  Lapata}]{liu-etal-2019-discourse}
Jiangming Liu, Shay~B. Cohen, and Mirella Lapata. 2019{\natexlab{a}}.
\newblock \href {https://doi.org/10.18653/v1/P19-1629} {Discourse
  representation parsing for sentences and documents}.
\newblock In \emph{Proceedings of the 57th Annual Meeting of the Association
  for Computational Linguistics}, pages 6248--6262, Florence, Italy.
  Association for Computational Linguistics.

\bibitem[{Liu et~al.(2019{\natexlab{b}})Liu, Cohen, and
  Lapata}]{liu-2019-discourse-representation}
Jiangming Liu, Shay~B. Cohen, and Mirella Lapata. 2019{\natexlab{b}}.
\newblock \href {https://doi.org/10.18653/v1/W19-1203} {Discourse
  representation structure parsing with recurrent neural networks and the
  transformer model}.
\newblock In \emph{Proceedings of the {IWCS} Shared Task on Semantic Parsing},
  Gothenburg, Sweden. Association for Computational Linguistics.

\bibitem[{Liu et~al.(2021)Liu, Cohen, and Lapata}]{liu-2021-generation}
Jiangming Liu, Shay~B. Cohen, and Mirella Lapata. 2021.
\newblock \href {https://www.aclweb.org/anthology/2021.naacl-main.35} {Text
  generation from discourse representation structures}.
\newblock In \emph{Proceedings of the 2021 Conference of the North American
  Chapter of the Association for Computational Linguistics: Human Language
  Technologies}, pages 397--415, Online. Association for Computational
  Linguistics.

\bibitem[{Liu et~al.(2020)Liu, Gu, Goyal, Li, Edunov, Ghazvininejad, Lewis, and
  Zettlemoyer}]{liu-etal-2020-mbart-denoising}
Yinhan Liu, Jiatao Gu, Naman Goyal, Xian Li, Sergey Edunov, Marjan
  Ghazvininejad, Mike Lewis, and Luke Zettlemoyer. 2020.
\newblock \href {https://doi.org/10.1162/tacl_a_00343} {Multilingual denoising
  pre-training for neural machine translation}.
\newblock \emph{Transactions of the Association for Computational Linguistics},
  8:726--742.

\bibitem[{Narayan and Gardent(2014)}]{narayan-gardent-2014-hybrid}
Shashi Narayan and Claire Gardent. 2014.
\newblock \href {https://doi.org/10.3115/v1/P14-1041} {Hybrid simplification
  using deep semantics and machine translation}.
\newblock In \emph{Proceedings of the 52nd Annual Meeting of the Association
  for Computational Linguistics (Volume 1: Long Papers)}, pages 435--445,
  Baltimore, Maryland. Association for Computational Linguistics.

\bibitem[{Papineni et~al.(2002)Papineni, Roukos, Ward, and jing Zhu}]{2002bleu}
Kishore Papineni, Salim Roukos, Todd Ward, and Wei jing Zhu. 2002.
\newblock {BLEU}: a method for automatic evaluation of machine translation.
\newblock In \emph{Proceedings of the 40th Annual Meetings of the ACL}, pages
  311--318.

\bibitem[{Poelman et~al.(2022)Poelman, van Noord, and Bos}]{wessel2022}
Wessel Poelman, Rik van Noord, and Johan Bos. 2022.
\newblock \href {https://aclanthology.org/2022.coling-1.367} {Transparent
  semantic parsing with {U}niversal {D}ependencies using graph
  transformations}.
\newblock In \emph{Proceedings of the 29th International Conference on
  Computational Linguistics}, pages 4186--4192, Gyeongju, Republic of Korea.
  International Committee on Computational Linguistics.

\bibitem[{Qiu et~al.(2020)Qiu, Sun, Xu, Shao, Dai, and Huang}]{xipeng2020}
Xipeng Qiu, Tianxiang Sun, Yige Xu, Yunfan Shao, Ning Dai, and Xuanjing Huang.
  2020.
\newblock Pre-trained models for natural language processing: A survey.
\newblock \emph{Science China Technological Sciences}, page 1872–1897.

\bibitem[{Rei et~al.(2020)Rei, Stewart, Farinha, and
  Lavie}]{rei-etal-2020-comet}
Ricardo Rei, Craig Stewart, Ana~C Farinha, and Alon Lavie. 2020.
\newblock \href {https://doi.org/10.18653/v1/2020.emnlp-main.213} {{COMET}: A
  neural framework for {MT} evaluation}.
\newblock In \emph{Proceedings of the 2020 Conference on Empirical Methods in
  Natural Language Processing (EMNLP)}, pages 2685--2702, Online. Association
  for Computational Linguistics.

\bibitem[{Tang et~al.(2022)Tang, Li, Zhao, and Wen}]{tang2022mvp}
Tianyi Tang, Junyi Li, Wayne~Xin Zhao, and Ji-Rong Wen. 2022.
\newblock \href {http://arxiv.org/abs/2206.12131} {{MVP}: Multi-task supervised
  pre-training for natural language generation}.
\newblock \emph{arXiv preprint, arXiv: 2206.12131v1}.

\bibitem[{Tang et~al.(2020)Tang, Tran, Li, Chen, Goyal, Chaudhary, Gu, and
  Fan}]{tang2020multilingual}
Yuqing Tang, Chau Tran, Xian Li, Peng-Jen Chen, Naman Goyal, Vishrav Chaudhary,
  Jiatao Gu, and Angela Fan. 2020.
\newblock \href {http://arxiv.org/abs/2008.00401} {Multilingual translation
  with extensible multilingual pretraining and finetuning}.
\newblock \emph{arXiv preprint, arXiv: 2008.00401v1}.

\bibitem[{van Noord et~al.(2018)van Noord, Abzianidze, Toral, and
  Bos}]{van-2018-exploring}
Rik van Noord, Lasha Abzianidze, Antonio Toral, and Johan Bos. 2018.
\newblock \href {https://doi.org/10.1162/tacl_a_00241} {Exploring neural
  methods for parsing discourse representation structures}.
\newblock \emph{Transactions of the Association for Computational Linguistics},
  6:619--633.

\bibitem[{van Noord et~al.(2019)van Noord, Toral, and
  Bos}]{van-2019-linguistic}
Rik van Noord, Antonio Toral, and Johan Bos. 2019.
\newblock \href {https://doi.org/10.18653/v1/W19-0504} {Linguistic information
  in neural semantic parsing with multiple encoders}.
\newblock In \emph{Proceedings of the 13th International Conference on
  Computational Semantics - Short Papers}, pages 24--31, Gothenburg, Sweden.
  Association for Computational Linguistics.

\bibitem[{van Noord et~al.(2020{\natexlab{a}})van Noord, Toral, and
  Bos}]{van-2020-character}
Rik van Noord, Antonio Toral, and Johan Bos. 2020{\natexlab{a}}.
\newblock \href {https://doi.org/10.18653/v1/2020.emnlp-main.371}
  {Character-level representations improve {DRS}-based semantic parsing even in
  the age of {BERT}}.
\newblock In \emph{Proceedings of the 2020 Conference on Empirical Methods in
  Natural Language Processing (EMNLP)}, pages 4587--4603, Online. Association
  for Computational Linguistics.

\bibitem[{van Noord et~al.(2020{\natexlab{b}})van Noord, Toral, and
  Bos}]{van-noord-etal-2020-character}
Rik van Noord, Antonio Toral, and Johan Bos. 2020{\natexlab{b}}.
\newblock \href {https://doi.org/10.18653/v1/2020.emnlp-main.371}
  {Character-level representations improve {DRS}-based semantic parsing even in
  the age of {BERT}}.
\newblock In \emph{Proceedings of the 2020 Conference on Empirical Methods in
  Natural Language Processing (EMNLP)}, pages 4587--4603, Online. Association
  for Computational Linguistics.

\bibitem[{Vaswani et~al.(2017)Vaswani, Shazeer, Parmar, Uszkoreit, Jones,
  Gomez, Kaiser, and Polosukhin}]{vaswani2017attention}
Ashish Vaswani, Noam Shazeer, Niki Parmar, Jakob Uszkoreit, Llion Jones,
  Aidan~N Gomez, \L~ukasz Kaiser, and Illia Polosukhin. 2017.
\newblock Attention is all you need.
\newblock In \emph{Advances in Neural Information Processing Systems},
  volume~30. Curran Associates, Inc.

\bibitem[{Wang et~al.(2021{\natexlab{a}})Wang, van Noord, Bisazza, and
  Bos}]{wang-2021-evaluating}
Chunliu Wang, Rik van Noord, Arianna Bisazza, and Johan Bos.
  2021{\natexlab{a}}.
\newblock \href {https://doi.org/10.18653/v1/2021.gem-1.8} {Evaluating text
  generation from discourse representation structures}.
\newblock In \emph{Proceedings of the 1st Workshop on Natural Language
  Generation, Evaluation, and Metrics (GEM 2021)}, pages 73--83, Online.
  Association for Computational Linguistics.

\bibitem[{Wang et~al.(2021{\natexlab{b}})Wang, van Noord, Bisazza, and
  Bos}]{wang-etal-2021-input}
Chunliu Wang, Rik van Noord, Arianna Bisazza, and Johan Bos.
  2021{\natexlab{b}}.
\newblock \href {https://doi.org/10.18653/v1/2021.acl-short.97} {Input
  representations for parsing discourse representation structures: Comparing
  {E}nglish with {C}hinese}.
\newblock In \emph{Proceedings of the 59th Annual Meeting of the Association
  for Computational Linguistics and the 11th International Joint Conference on
  Natural Language Processing (Volume 2: Short Papers)}, pages 767--775,
  Online. Association for Computational Linguistics.

\bibitem[{Wang et~al.(2022)Wang, Xu, Fang, Liu, Sun, Xu, Zhu, and
  Zeng}]{wang-etal-2022-training}
Shuohang Wang, Yichong Xu, Yuwei Fang, Yang Liu, Siqi Sun, Ruochen Xu,
  Chenguang Zhu, and Michael Zeng. 2022.
\newblock Training data is more valuable than you think: A simple and effective
  method by retrieving from training data.
\newblock In \emph{Proceedings of the 60th Annual Meeting of the Association
  for Computational Linguistics (Volume 1: Long Papers)}, Online. Association
  for Computational Linguistics.

\bibitem[{Wolf et~al.(2020)Wolf, Debut, Sanh, Chaumond, Delangue, Moi, Cistac,
  Rault, Louf, Funtowicz, Davison, Shleifer, von Platen, Ma, Jernite, Plu, Xu,
  Le~Scao, Gugger, Drame, Lhoest, and Rush}]{wolf-etal-2020-transformers}
Thomas Wolf, Lysandre Debut, Victor Sanh, Julien Chaumond, Clement Delangue,
  Anthony Moi, Pierric Cistac, Tim Rault, Remi Louf, Morgan Funtowicz, Joe
  Davison, Sam Shleifer, Patrick von Platen, Clara Ma, Yacine Jernite, Julien
  Plu, Canwen Xu, Teven Le~Scao, Sylvain Gugger, Mariama Drame, Quentin Lhoest,
  and Alexander Rush. 2020.
\newblock \href {https://doi.org/10.18653/v1/2020.emnlp-demos.6} {Transformers:
  State-of-the-art natural language processing}.
\newblock In \emph{Proceedings of the 2020 Conference on Empirical Methods in
  Natural Language Processing: System Demonstrations}, pages 38--45, Online.
  Association for Computational Linguistics.

\bibitem[{Xue et~al.(2021)Xue, Constant, Roberts, Kale, Al-Rfou, Siddhant,
  Barua, and Raffel}]{xue-etal-2021-mt5}
Linting Xue, Noah Constant, Adam Roberts, Mihir Kale, Rami Al-Rfou, Aditya
  Siddhant, Aditya Barua, and Colin Raffel. 2021.
\newblock \href {https://doi.org/10.18653/v1/2021.naacl-main.41} {m{T}5: A
  massively multilingual pre-trained text-to-text transformer}.
\newblock In \emph{Proceedings of the 2021 Conference of the North American
  Chapter of the Association for Computational Linguistics: Human Language
  Technologies}, pages 483--498, Online. Association for Computational
  Linguistics.

\end{thebibliography}
\bibliographystyle{acl_natbib}

\clearpage

\appendix
\onecolumn

\section{Appendix}
\label{sec:appendix}

\subsection{Human Evaluation}
Human evaluation was performed on a preliminary version of the models to assess correlation with the automatic metrics we planned to use on all larger-scale experiments.
We adopt ROSE~\citep{wang-2021-evaluating}, a human evaluation method that covers three dimensions: \emph{semantics}, \emph{grammaticality} and \emph{phenomenon}, to assess the performance of models' outputs in the generation task.
Since we are not investigating a particular linguistic phenomenon, we focus on the first two dimensions only: meaning preservation (whether the generated text has the same meaning as the gold text) and grammaticality (whether the generated text has no grammatical errors). We ask two experts with a linguistic doctorate degree to rate the generated texts with \{0: No, 1: Yes\} on these two dimensions. To reduce the annotation load, we exclude all outputs that are identical to the corresponding references, and then randomly select 100 samples for each language.

\paragraph{Evaluation Results} 

Table~\ref{table:human-results} shows that about 26\% of the generated sentences in languages other than English correspond to the references, while this rate is about 50\% for English it reaches around 50\%, due to the larger datasets.
While the training data for German and Italian also far exceeds that of Dutch, the evaluation results are very close (including automatic evaluation) for these three languages, suggesting that the models do benefit from cross-lingual knowledge transfer.

\begin{table}[!h]
\centering
\footnotesize
\setlength{\tabcolsep}{5pt}
\begin{tabular}{ccccc}
\toprule
\textbf{Lang} & \textbf{Perfect} & \textbf{Semantics} & \textbf{Grammaticality} & \textbf{Overall} \\
\hline
  \textbf{EN} & 49.3 & 87.0 & 90.0 & 83.0 \\
  \textbf{DE} & 26.4 & 54.0 & 85.0 & 45.0 \\
  \textbf{IT} & 27.2 & 51.0 & 70.0 & 38.0\\
  \textbf{NL} & 26.3 & 51.0 & 74.0 & 45.0\\
\bottomrule
\end{tabular}
\caption{Human evaluation results (\%). 
\texttt{Perfect} indicates the ratio of generated sentences which correspond exactly to the human references; (ii) \texttt{Overall} indicates the ratio of cases rated 1 for both semantics and grammaticality.}
\label{table:human-results}
\end{table}



\end{document}